\documentclass[sigconf]{acmart}
\usepackage{soul}
\usepackage{xcolor}
\usepackage{amsmath}
\usepackage{amsfonts}
\usepackage{cleveref}
\usepackage{algorithm}
\usepackage{algpseudocode}
\usepackage[list=true]{subcaption}

\AtBeginDocument{%
  \providecommand\BibTeX{{%
    \normalfont B\kern-0.5em{\scshape i\kern-0.25em b}\kern-0.8em\TeX}}}

\copyrightyear{2022} 
\acmYear{2022} 
\setcopyright{acmlicensed}\acmConference[GECCO '22 Companion]{Genetic and Evolutionary Computation Conference Companion}{July 9--13, 2022}{Boston, MA, USA}
\acmBooktitle{Genetic and Evolutionary Computation Conference Companion (GECCO '22 Companion), July 9--13, 2022, Boston, MA, USA}
\acmPrice{15.00}
\acmDOI{10.1145/3520304.3533966}
\acmISBN{978-1-4503-9268-6/22/07}

\acmConference[GECCO '22]{GECCO '22: The Genetic and Evolutionary Computation Conference}{July 09--13, 2022}{Boston, MA}
\begin{document}

\title{Towards Explainable Metaheuristic: Mining Surrogate Fitness Models for Importance of Variables}

\author{Manjinder Singh}
\email{manjinder.singh1@stir.ac.uk}
\orcid{0000-0003-4720-3473}

\affiliation{%
  \institution{University of Stirling}
  \city{Stirling}
  \country{UK}
}

\author{Alexander E.I. Brownlee}
\email{alexander.brownlee@stir.ac.uk}
\orcid{0000-0003-2892-5059}

\affiliation{%
  \institution{University of Stirling}
  \city{Stirling}
  \country{UK}
}

\author{David Cairns}
\email{david.cairns@stir.ac.uk}
\orcid{0000-0002-0246-3821}

\affiliation{%
  \institution{University of Stirling}
  \city{Stirling}
  \country{UK}
}

\renewcommand{\shortauthors}{M. Singh et al.}


\begin{abstract}
Metaheuristic search algorithms look for solutions that either maximise or minimise a set of objectives, such as cost or performance. However most real-world optimisation problems consist of nonlinear problems with complex constraints and conflicting objectives.

The process by which a GA arrives at a solution remains largely unexplained to the end-user. A poorly understood solution will dent the confidence a user has in the arrived at solution. We propose that investigation of the variables that strongly influence solution quality and their relationship would be a step toward providing an explanation of the near-optimal solution presented by a metaheuristic. 

Through the use of four benchmark problems we use the population data generated by a Genetic Algorithm (GA) to train a surrogate model, and investigate the learning of the search space by the surrogate model. We compare what the surrogate has learned after being trained on population data generated after the first generation and contrast this with a surrogate model trained on the population data from all generations. 

We show that the surrogate model picks out key characteristics of the problem as it is trained on population data from each generation. Through mining the surrogate model we can build a picture of the learning process of a GA, and thus an explanation of the solution presented by the GA. The aim being to build trust and confidence in the end-user about the solution presented by the GA, and encourage adoption of the model.

\end{abstract}




\maketitle

\section{Introduction}
The growing ubiquity of machine learning (ML) and Artificial Intelligence (AI) models \cite{olhede_growing_2018} to automate decision making in our day-to-day life has brought with it ethics concerns\cite{yapo_ethical_2018}, and a lack of trust from those users directly impacted by the automated decision making \cite{toreini_relationship_2020,aoki_experimental_2020}. 
The interest in explainability is not a new one \cite{david_explanation_1993}, however the resurgence in interest is being driven by users demanding an explanation of {\itshape why} a particular decision was reached. This demand for an explanation has also been enshrined in European Union (EU) law when the right to explanation was included in the General Data Protection Regulation (GDPR) \cite{goodman_european_2017} as a recognition of the rise in importance of AI systems in automated decision making.

The majority of AI models in use today can be thought of as {\itshape ''black-boxes''}, with many complex layers consisting of nonlinear transformations \cite{lecun_deep_2015}. It can be difficult for end users to understand the decision making process and hence, already, trust in the final output is eroded. This has led to decisions being made that were biased
\cite{lambrecht_algorithmic_2019}, and some that have led to real harm being done to people due to a lack of transparency and explainability of these models \cite{rudin_stop_2019}.

A large area within AI is metaheuristic search algorithms applied to optimisation problems. These algorithms look for solutions that either maximise or minimise a set of objectives, such as cost or performance. However most real-world optimisation problems consist of nonlinear problems with complex constraints and conflicting objectives. These algorithms find applications in many areas, including optimisation of transportation for positive environmental impact \cite{brownlee_fuzzy_2018}, and healthcare \cite{ochoa_multi-objective_2020}.

Metaheuristic algorithms are not problem dependent, which means we can use them for a variety of optimisation problems. Many of these algorithms are inspired by nature. A Genetic Algorithm (GA) takes its inspiration from evolution. To find an optimal or near-optimal solution for a problem, the GA is used to optimise a problem by initially creating many random solutions, populations, and then iteratively updating these solutions through selection, mutation and crossover, to reach a {\itshape good enough} solution or the most optimal solution possible. A GA performs well in combinatorial optimisation where we have a large search space \cite{anderson_genetic_1994}. However, despite recent advances in theoretical understanding of GAs \cite{malan_survey_2021, lehre_theoretical_2017}, the process by which a GA arrives at a solution remains largely unexplained to the end-user. Herein lies the problem, firstly a poorly understood solution will dent the confidence a user has in the arrived at solution. Secondly during our problem definition we may have excluded some critical criteria. This may be of particular relevance in a scheduling optimisation problem, where we may focus on only a distance or financial cost constrain and so fail to take into account personal preferences and/or convenience for the end-user. Thirdly, a GA follows a random process, which can lead to noise in the presented solution \cite{di_pietro_applying_2004}. In this case it would be useful to know the characteristics of this solution, and be able to tease out this noise without impacting the solution quality. Metaheuristic algorithms in general are adept at finding {\itshape shortcuts} in a problem definition \cite{lehman_surprising_2020}. Therefore it is useful to know whether the solution presented genuinely solves the problem or if the algorithm merely found a {\itshape loophole} in the problem definition. As demonstrated, for the Vehicle Routing Problem (VRP)\cite{arnold_what_2019}, being able to explain the characteristics that constitute a near-optimal or sub-optimal solution also goes a long way towards designing a more efficient algorithm.

Taking all of the above points into consideration we can see that there are many points at which we can lose a user's trust when designing a system. This loss of trust would ultimately lead to lack of adoption of the system, and hence, wasted resources and suspicion of any systems recommended in the future. % The search processes that are at the core of metaheuristic algorithms are random and non-interpretable to humans, and can be thought of as a {\itshape ''black-box''} \cite{lecun_deep_2015}. A metaheuristic problem formally defined using a fitness function. This fitness function is used to evaluate the solutions, and is difficult to interrogate directly in order to extract meaningful explanations.

In this paper we will outline an approach to identify the cardinal characteristics of metaheuristic derived solutions. We do this for a series of binary-encoded benchmark problems, by way of identifying which variables have the greatest influence on solution quality. Our approach will use {\itshape surrogate fitness functions} \cite{brownlee_constrained_2015,jin_surrogate-assisted_2011}. Usually these are applied where the fitness function is costly; in parallel to the optimisation process we train a computationally cheap model, and the majority of calls to the costly fitness function are replaced by a call to the surrogate model. An additional benefit of this process is that the surrogate is an explicit representation of what the metaheuristic has learned about the search space and more crucially what comprises a {\itshape good} solution. In our study, we take a high fitness solution for each of the benchmark problems and probe it using the surrogate model to identify which variables are of greatest importance and to test their impact on fitness. The present paper extends the approach initially proposed in \cite{bramer_towards_2021}, by investigating the insights that might be gained at different stages of the algorithm's run, rather than just the final population.

In {\itshape Section \ref{related-work} } we will review related work around trust and explanation of metaheuristic optimisation, before going on to focus on the role of surrogate models in optimisation problems, and our use of them to explain metaheuristic algorithms. In {\itshape Section \ref{surrogate-models}} we describe our approach to mining the surrogate model to explain solution quality. In {\itshape Section \ref{methodology}} we outline our approach to mining the surrogate model to extract information about variable importance of an optimal solution. In {\itshape Section \ref{experiments}} we demonstrate our methodology using four benchmark problems, and in {\itshape Section \ref{results} } we present our results and outline our conclusions and future work in {\itshape Section \ref{conclusion}}.

\section{Related Work}\label{related-work}
Explainability in AI is not a new research topic, however, research conducted in the early 90's \cite{david_explanation_1993} tapered off for many decades. The recent growth and interest in this area is gaining momentum, mainly driven by a lack of trust and uptake by users of AI systems. The resurgence in this topic follows the thinking that making AI systems more transparent and explainable will engender end-user trust in the AI systems and hopefully lead to greater uptake and use of these systems \cite{miller_explanation_2017}.

There is, however, a gap in research centered around explainability of metaheuristic algorithms. The most relevant research within this area being innovization outlined in a paper by Deb et al. \cite{deb_integrated_2014} \cite{deb_innovization_2006}, which proposed {\itshape ``innovization''} to generate problem based knowledge alongside the normally generated near-optimal solutions, through identification of common principles among Pareto-optimal solutions for multi-objective optimisation problems. The impetus behind this being that most optimisation techniques adopted are used to surrender a single or small selection of optimal solutions, their research built on this to allow these optimisation techniques to find additional problem based knowledge in parallel to the generated optimal solution(s) via the innovization operation. By taking a generated set of {\itshape high-performing} trade-off solutions and identifying common principles hidden within them. The idea is that the main common thread amongst this set of {\itshape high-performing} solutions will represent properties that ensure Pareto-Optimality and by extension are valuable properties related to the problem globally. 

More recently proposed by Urquhart et al. \cite{urquhart_increasing_2019} is an application of Map-Elites to increase trust in metaheuristic algorithms. This paper aims to address the criticism that end-users have no role in the construction of the end solution. The authors applied Map-Elites, which allows for construction of a {\itshape high-performance} solution which is mapped onto a set of solutions defined by the user, such as cost or time. The solutions being generated by mutation and recombination, with each solution being assigned a {\itshape bin} within the solution space, with solutions that are assigned to an already occupied bin being either rejected or accepted based on a comparison of their relative fitness, with higher ranked fitness solutions being accepted, so only the highest fitness being assigned and retained within any particular {\itshape bin}. Ultimately the end-user is responsible for choosing the preferred solution, this ensures that input from the user is taken into account and they have a sense of ownership and a greater level of trust in the proposed solution. Map-Elites therefore provide a way to filter the solution space, and provides a set of solutions for the user, from which they can select the one most applicable to them and their needs. This has the benefit of increasing trust in the solution selected, because the user is provided an opening to the process and able to have some measure of influence as to what constitutes a {\itshape good} solution.

Gaier et al. \cite{gaier_data-efficient_2017} proposed a hybrid approach by using Map-Elites alongside a surrogate model to add efficiency to the Map-Elites process. They proposed reducing the need for the large number of checks normally required for Map-Elites. The proposed solution, Surrogate-assisted illumination (SAIL), aims to achieve this by way of integrating an approximation model (surrogate) alongside intelligent sampling of the fitness function. As with Map-Elites the search space is partitioned into {\itshape bins} each of which holds a map with a different layout of feature values. Firstly a surrogate is constructed based on an initial population of possible solutions including their fitness scores. Map-Elites is then used to produce solutions to maximise the fitness function, and generating an acquisition map. Thereafter, new solutions are sampled from this map and additional observations are used to iteratively improve the model. With the aim of looping through this process to generate increasingly better solutions with higher fitness functions. The performance predictions then being used by Map-Elites in place of the original fitness function to generate a prediction map of near optimal representations.

In contrast to the above approaches we do not aim to summarise or present a set of solutions, rather, we aim to explain a single solution chosen by the metaheuristic algorithm.

\section{Surrogate Models}\label{surrogate-models}
Evolutionary Algorithms (EA) aim to minimise or maximise a given function {\itshape f(X)}. This is done by initially generating random solutions and evaluating them against a {\itshape fitness function}, and then generating new solutions, usually biased towards solutions with higher fitness values. The fitness function is often the most costly operation in the optimisation process, and to reduce this cost we can replace calls to the {\itshape fitness function} with calls to a much cheaper model. This cheaper model, {\itshape surrogate} \cite{brownlee_metaheuristic_2015,jin_surrogate-assisted_2011,jin_comprehensive_2005, bramer_towards_2021}, also represents an explicit model of the population. We propose mining this model to capture the sensitivity of the {\itshape fitness function} to the problem variables. The premise being that the surrogate model is biased by the population of the EA, and therefore contains another view of the algorithms understanding of the problem. 

As in \cite{bramer_towards_2021}, in the work presented here, the EA alternates between using the surrogate to evaluate solutions and the true fitness function. The surrogate model is re-trained each time new evaluations are carried out by the true fitness function, it is re-trained on the current and all previous populations at alternating intervals. So, we alternate between using the true fitness function and the surrogate at regular intervals. The surrogate is implemented by a Support Vector Regression (SVR) model, provided by scikit-learn (version 1.0.2). We use the default parameter settings for the SVR model. The model features are simply the problem variables {\itshape X} and the target for prediction is fitness {\itshape f(X)}. We hypothesise that the surrogate model will retain some of the key properties from the original fitness function, such as regions of high fitness. The surrogate model is able to evaluate the solution at a much cheaper cost compared to the real fitness function. The two benefits of this approach being that firstly the surrogate model can evaluate the potential solutions in a more timely manner and, secondly can give us an insight into the algorithms understanding of the search space.

The benefit of this approach is that the surrogate model, being an explicit model of the population, can be mined to gain insights into the learning of the EA of the search space. This approach has previously been demonstrated when looking at Estimation of Distribution Algorithms (EDAs), which sample probabilistic models after first constructing these models to reflect the distribution of high fitness solutions. It has been demonstrated that information gained from EDAs can in many cases offer additional insight into the problem as the solutions presented by an EA.\cite{brownlee_using_2010, brownlee_application_2013,brownlee_mining_2016}. The usefulness of the information gleaned from a surrogate model is of course closely coupled to the problem being analysed. The surrogate model could be inherently interpretable (such as a linear regression model similar to \cite{brownlee_mining_2016} or a decision tree), or we could exploit existing XAI approaches such as mining the model through probing. We make use of the latter approach in the present paper.

\section{Methodology}\label{methodology}
To determine the rank and importance of individual variables for each of the benchmark problems, we will investigate the local sensitivity of the closest-to-optimal solution ({\itshape highest fitness}) found for each of the problems. We will take each of these {\itshape best solutions} and evaluate against our surrogate model, allowing us to determine the change in surrogate fitness due to mutating each variable in the solution. We will also set the sign of the {\itshape ``importance''} measure to be negative if the corresponding bit in the starting seed solution was 1: providing an indication of the direction of the relationship between that variable and the fitness. This gives us a measure of the variables importance and its correlation with fitness, in the neighbourhood of the {\itshape high-fitness} solution.

We aim to highlight the variables that impact either negatively or positively on solution quality. It should be noted that this approach is applied to benchmark problems for demonstration purposes only, where the simple problem definition means that the explanations can be compared against known expectations, and gives a starting point for further research in this area for real-world problems. The surrogate is trained on data derived from the real fitness function, hence, it learns the local search space from, and is biased towards, the best solutions from the metaheuristic. In this way the surrogate reflects the understanding of the problem as learned by the algorithm. 

We will compare the surrogate model trained on only the first generation against a surrogate trained on all populations over the run of 100 generations. Because the surrogate is biased toward the best solution, we aim to visualise the surrogates understanding of the problem between the first generation and over all generations.

A formal definition of our approach is shown in \Cref{algorithm:mining-algorithm}.

\begin{algorithm}
    \begin{algorithmic}
    \State \textbf{In:} \textit{$x=(x_0 \ldots x_n), x_i=\{0,1\}$}, near-optimal \textit{seed} solution found by GA
    \State \textbf{In:} $S(X)\rightarrow f$, surrogate fitness function to estimate fitness $f$ of a solution $X$
    \State \textbf{Out:} $C=(c_0 \ldots c_n), c_i \in \mathbb{R}$, absolute change to surrogate fitness for each variable in $x$
    \State $C \leftarrow \emptyset$; \;
    \State $f_{org} \leftarrow S(x)$ \Comment{surrogate fitness of solution}
    \For{for each variable $x_i$}){$i = 0$ to $n-1$} 
        \State $x_i \leftarrow (x_i+1) \bmod{} 2$ \Comment{flip variable $x_i$}
        \State $\hat{f_i} \leftarrow S(x_i)$  \Comment{surrogate fitness of mutated solution}
        \If{$x_i=1$}
            $c_i = -1 * \hat{f_i}$ \Comment{Change in surrogate fitness, optimum should have a 1}
        \Else
            $c_i = \hat{f_i}$ \Comment{Change in surrogate fitness, optimum should have a 0}
        \EndIf
  	    \State $C \leftarrow \lvert c_i \rvert $ \Comment{add to list}
    \EndFor
 \end{algorithmic}
 \caption{Probing variables in a solution with respect to the surrogate fitness function}
  \label{algorithm:mining-algorithm}
\end{algorithm}

\section{Experiments}\label{experiments}
We will focus on four well-known bit-string encoded benchmark functions ({\itshape 1D Checkerboard, 2D Checkerboard, Trap5 and MAXSAT}). 

The experiment setup is shown in \Cref{gatable}. Where {\itshape P} is the population size, {\itshape n} the length of the bit-string, problem-size, {\itshape maxGen} is the maximum number of generations, {\itshape mutRate} the mutation rate. The Selection operator used is {\itshape Tournament} with a tournament size of 5. The {\itshape Crossover} operator is Uniform. We have made little effort to tune the algorithm beyond simple empirical exploration of these parameters, because the focus of this study is on explaining the results rather than maximal algorithm performance.

\begin{table}[hbt]
\centering
\begin{tabular}{@{}cccccc@{}}
\toprule
P   & n  & maxGen & mutRate & Selection  & Crossover      \\ \midrule
100 & 100 & 100    & 0.01   & Tournament(5) & Uniform \\ \bottomrule
\end{tabular}
\caption{Experiment Setup}
\label{gatable}
\end{table}

\subsection{Benchmark Problems}
Our experiment focused on four benchmark functions using a bit-string representation. The four functions were chosen to contrast uni-variate and multi-variate problems. This will allow us to investigate variations in the importance attached to each variable and model's handling of the presence, or absence, of interactions.
\subsubsection{1D Checkerboard}

The objective of the 1D-Checkerboard problem \cite{Baluja-1997-16482} is to realise a checkerboard pattern with alternating 1s and 0s. The function scores the chromosome based on the sum of adjacent variables that do not share the same value. A formal description of the function is shown in \Cref{1d}.

\begin{equation}
    f(x) = \sum_{i=0}^{l-2}\begin{Bmatrix}
 1, & x_i\neq x_{i+1}\\ 
 0,& x_i =  x_{i+1}
\end{Bmatrix}
\label{1d}
\end{equation}

\subsubsection{2D Checkerboard}
The 2D-Checkerboard problem \cite{Baluja-1997-16482, larranaga_review_2002} introduces bivariate interactions into the problem. While these interactions are weighted equally in the fitness function, implicitly those in the centre of the grid have greater impact on fitness. A solution represents the rows of a {\itshape s} x {\itshape s} grid concatenated into one string, the objective being to realsie a grid with a checkerboard pattern of alternating 1s and 0s. A formal description of the function is shown in \Cref{2d}

\begin{equation}
    f(x) = 4(s-2)^{2} -\sum _{i=2}^{s-1} \sum _{j=2}^{s-1}\begin{Bmatrix}
    \delta \left(x_{ij} ,x_{i-1j} \right)+\delta \left(x_{ij} ,x_{i+1j} \right)\\
    +\delta \left(x_{ij} ,x_{ij-1} \right)+\delta \left(x_{ij} ,x_{ij+1} \right)
    \end{Bmatrix}
\label{2d}
\end{equation}

\subsubsection{Trap5}

The Trap-5 problem is designed to be intentionally deceptive \cite{deb_sufficient_1994}, by rewarding steps towards local optima in small groups of variables in order to force, or deceive, an EA away from the global optimum. This is of particular concern in algorithms that do not consider interactions between variables\cite{goldberg1992massive, sandy2009}. The problem construction consists of bit-strings partitioned into blocks with their fitness scored separately. A formal description of the function is shown in \Cref{trap51,trap52}

\begin{subequations}
\begin{equation}
f(x)=\sum_{i=1}^{n/k}trap_k({x_{b_{i+1}}}+...+{x_{b_{i+k}}})
\label{trap51}
\end{equation}

\begin{equation}
trap_k(u)=\left\{ f_\text{high}\text{ if }u=k, f_\text{low}-u\frac{f_\text{low}}{k-1} \text{  otherwise}\right\}
\label{trap52}
\end{equation}
\end{subequations}

\subsubsection{MAXSAT}

The Maximum Satisfiability or MAXSAT problem \cite{johnson_approximation_1974, goos_genetic_1998, brownlee_solving_2007} attempts to find a set of values which maximises the number of satisfied clauses of a fixed predicate logic formula expressed in conjunctive normal form (CNF). It is known to be NP-complete in its general form. The MAXSAT is useful for modeling high order interactions, because each instance of the problem uses a known predefined structure. The problem construction involves a bit-string, where each bit encodes a predicate variable in the CNF formula. An individuals fitness is therefore just equal to the number of satisfied clauses.

\subsection{Experimental Procedure}
For each of the benchmark problems, we started with an initial population of 100 individuals and an Genetic Algorithm (GA) was used to determine a near-optimal solution. A problem size of 100, population size of 100, tournament selection with size 5, mutation rate of 0.01, crossover rate of 0.95 and was run over 100 generations. We used elitism, to guarantee that the single fittest solution was carried forward for each generation without mutation.

A normal run of a GA would consist of generating an initial population for which a fitness value is calculated and through selection, crossover and mutation a second population is generated. This process iterates for a predetermined number of generations, or until a solution with the possible fitness, as defined by our fitness function, is found, whichever occurs first. At the end of this run, the most optimal solution found is presented to the user. However, the solution presented is devoid of an explanation, as to how it was found nor why it is considered the fittest solution, other than it has the {\itshape best fitness function value}, which was determined by the algorithm designer.

We will attempt to provide an intuitive understanding of the solution generation process, with the aim of providing an explanation for the end-user as to how the solution was chosen. We will do this via construction of a surrogate model, using the population and fitness values for each generation of the GA run as training data for the surrogate. The aim in this experiment is not to generate a surrogate to speed up the fitness evaluation process, but rather to mine the surrogate for importance of variables. We will examine the surrogate model trained on only the population form the first generation and one trained on the the population data for all of the runs over 100 generations.

For each problem we used a Support Vector Regression (SVR) model, provided by scikit-learn (version 1.0.2), as the surrogate. We used the default hyper-parameters for this model, and made no attempt to tune any of the parameters, so as to enable reproducible experiments and focus on the learning of the search space by the surrogate model between the first and all generations of the GA run.

\section{Results and Discussion}\label{results}

\subsection{Surrogate Trained on First Generation Data}\label{firstData}

\begin{figure}[htb]
  \centering
  \includegraphics[width=\linewidth]{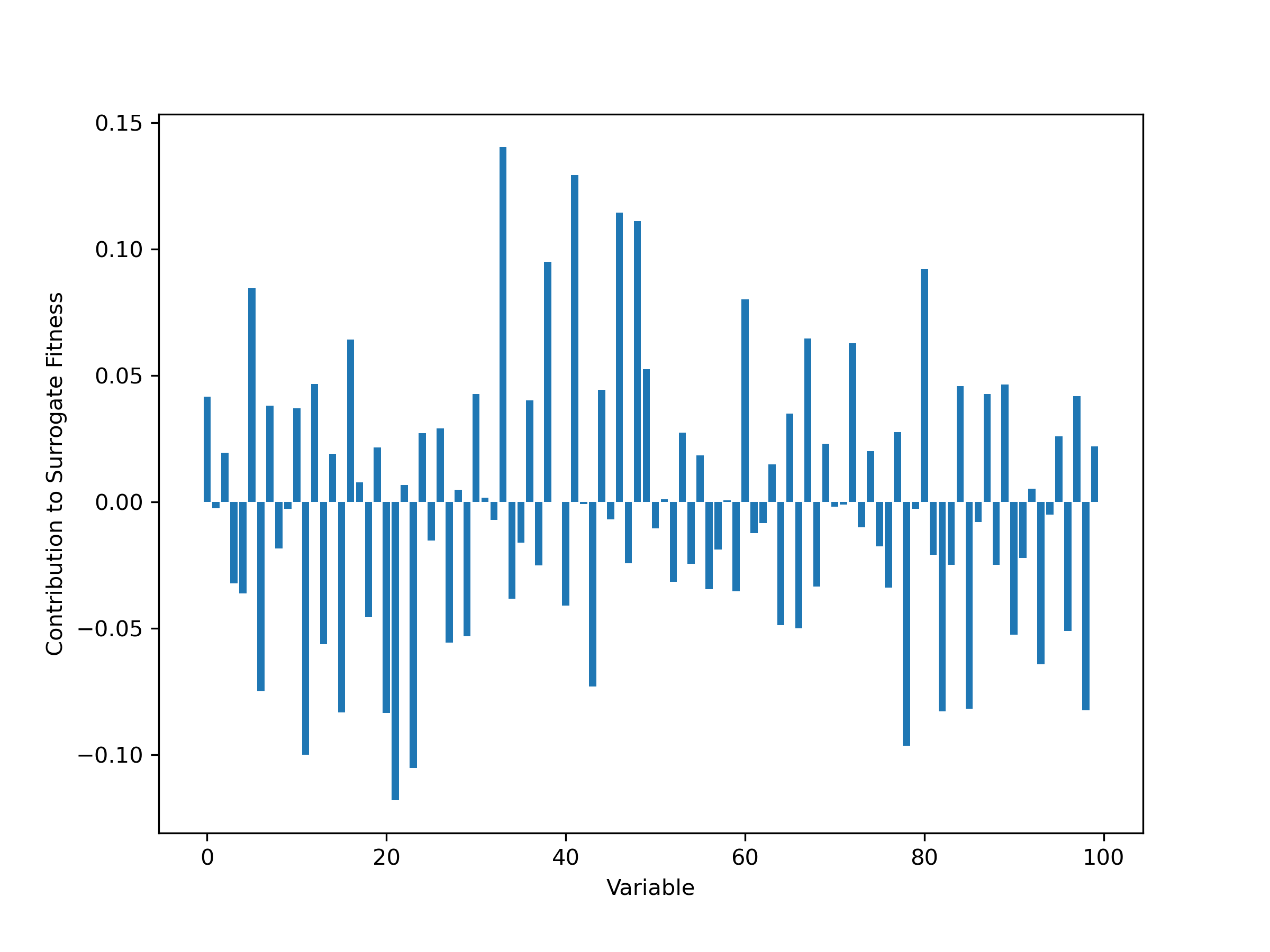}
  \caption{1D Checkerboard Contribution to Surrogate Fitness per Variable (First Generation)}
  \label{fig:1DCheck}
\end{figure}

\begin{figure}[htb]
  \centering
  \includegraphics[width=\linewidth]{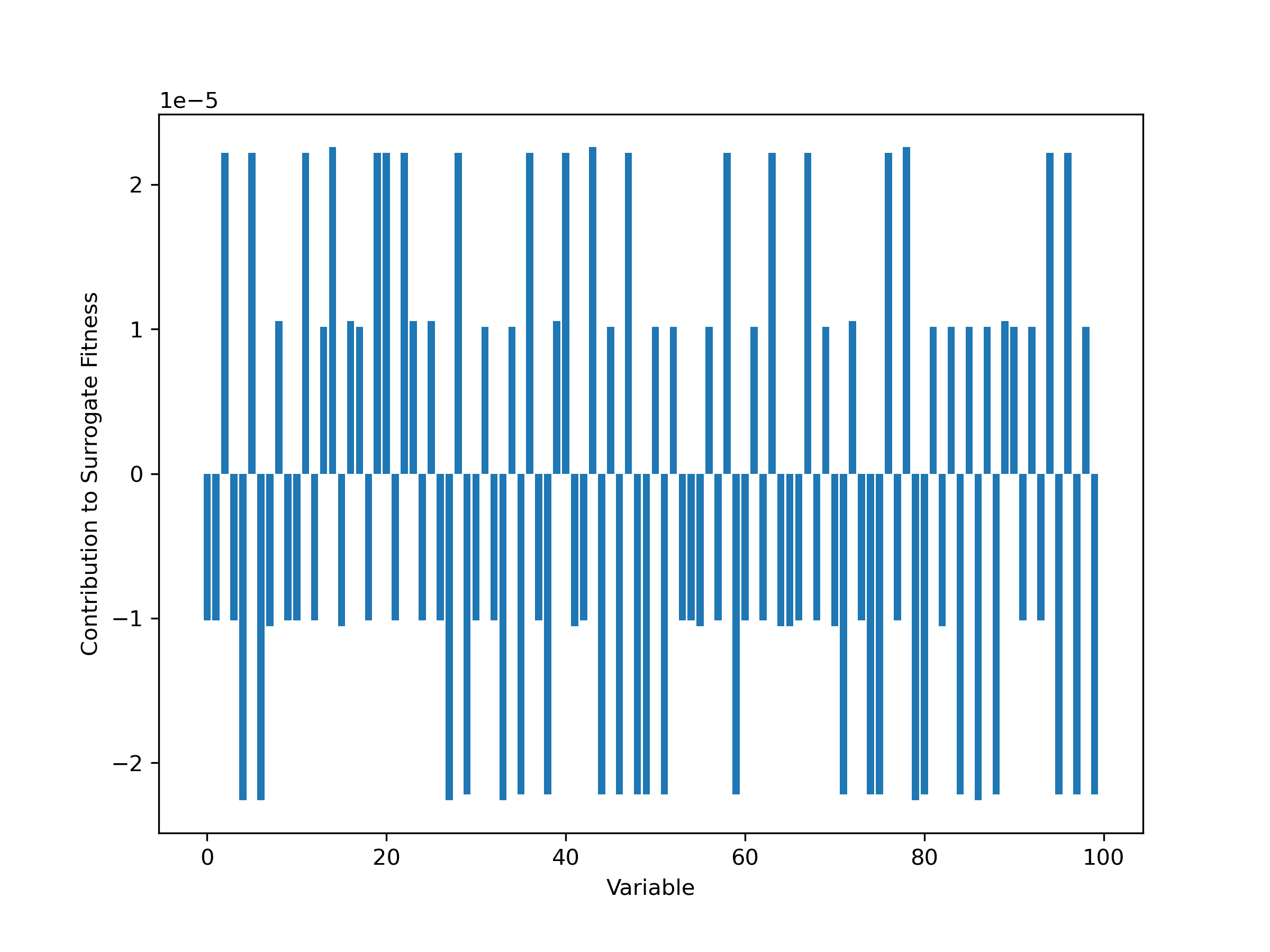}
  \caption{2D Checkerboard Contribution to Surrogate Fitness per Variable (First Generation)}
  \label{fig:2DCheck}
\end{figure}

\begin{figure}[htb]
  \centering
  \includegraphics[width=\linewidth]{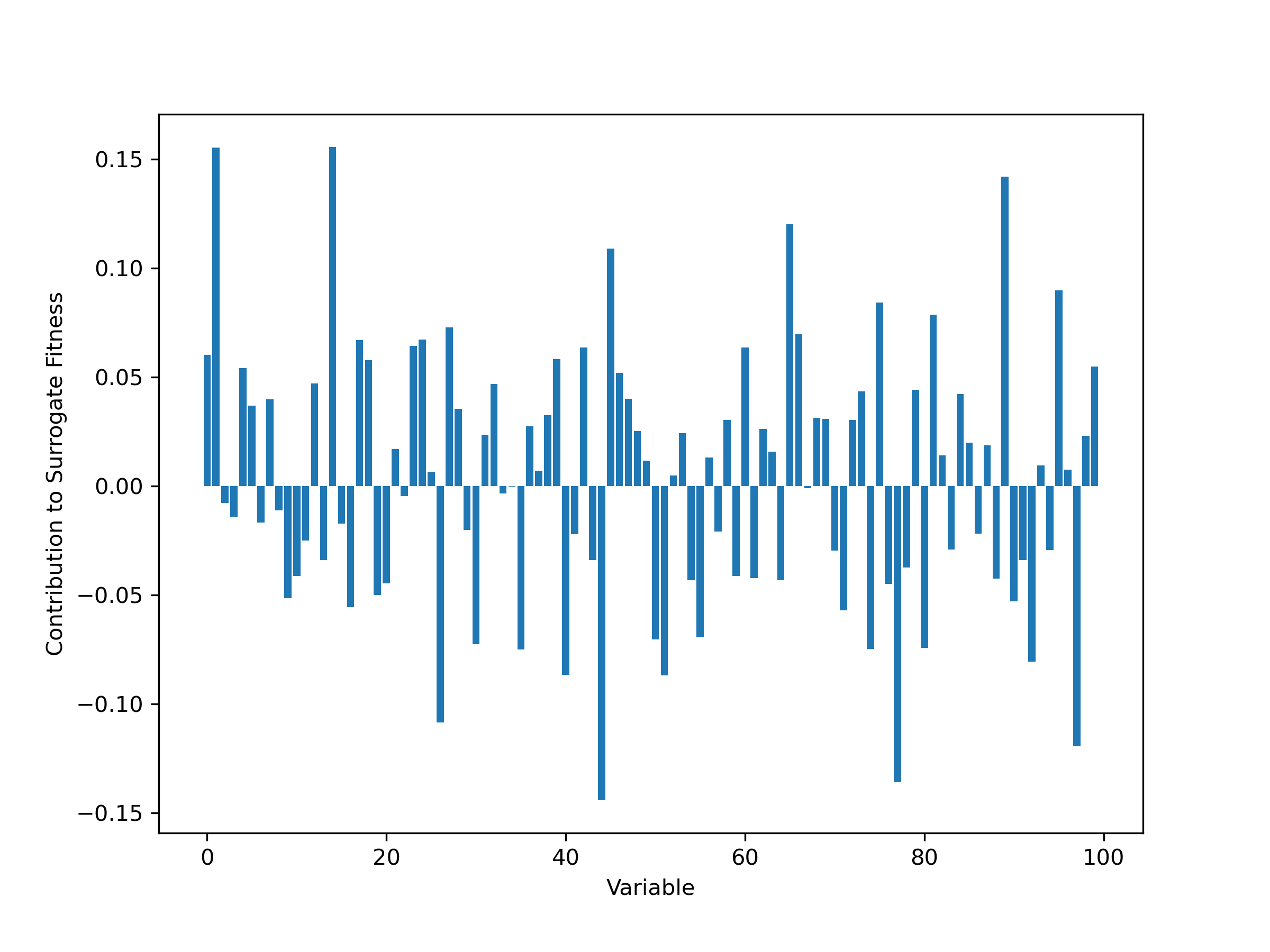}
  \caption{MAXSAT Contribution to Surrogate Fitness per Variable (First Generation)}
  \label{fig:MAXSAT}
\end{figure}

\begin{figure}[htb]
  \centering
  \includegraphics[width=\linewidth]{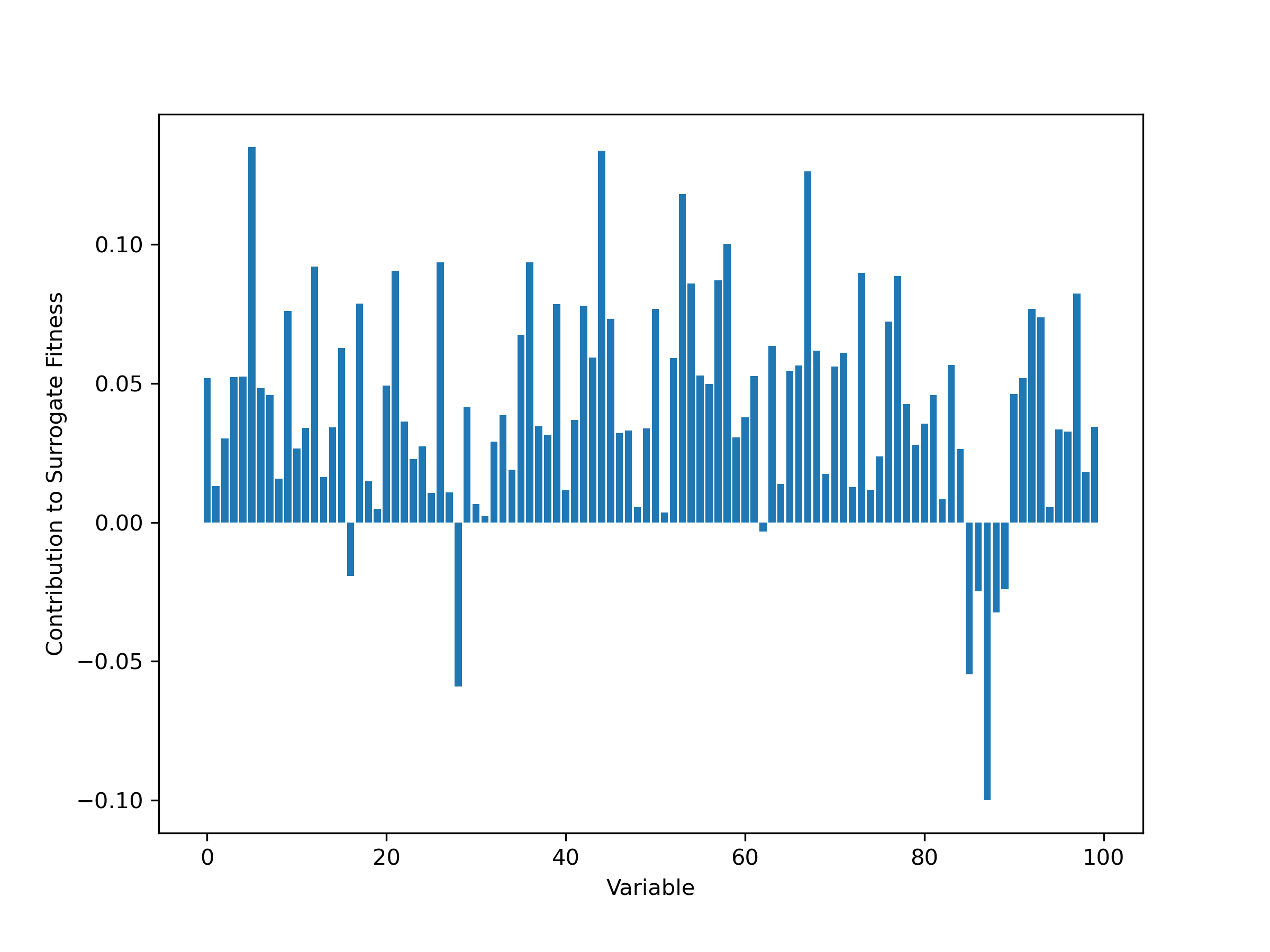}
  \caption{Trap5 Contribution to Surrogate Fitness per Variable (First Generation)}
  \label{fig:Trap5}
\end{figure}

In this section we will consider the results of mining a surrogate fitted to the first population of the algorithm run. Each plot in \Cref{fig:1DCheck,fig:2DCheck,fig:MAXSAT,fig:Trap5} shows the mean contribution to surrogate fitness for each variable. The positive values indicate that the optimal value was a 0 and negative values indicate that it was a 1. In \Cref{fig:1DCheck} we can observe that the model has not managed to grasp enough of the problem structure, there is a clear partial chain structure visible, however, it is difficult to see the relative contribution of each variable due to random noise. We would expect to see an equal contribution from all the variables. There is a similar picture with Trap5 (\Cref{fig:Trap5}) and MAXSAT (\Cref{fig:MAXSAT}). Although with the latter there should be some variation, as variables appear in different clauses and so have different contributions to overall fitness, there is only a small variation in the number of clauses a variable appears in, and not as much as suggested by the bar plot. For 2D checkerboard there is a clearer picture of the problem structure emerging, suggesting that some variable have twice the importance of others. This appears to be driven by the suboptimal solutions in the first generation; as we will see, the model fitted to all generations weights all variables similarly and shows clearer signs of the checkerboard pattern.

In there results, the model is trained on only the first population, at this point the GA is just starting an exploration of the search space, and so has not explored enough of the problem structure to generate meaningful data on which to train the surrogate model.

Our aim here was to set a starting point to which we can then compare the results as outlined in \Cref{allData}

\subsection{Surrogate Trained on All Generation Data}\label{allData}

\begin{figure}[htb]
  \centering
  \includegraphics[width=\linewidth]{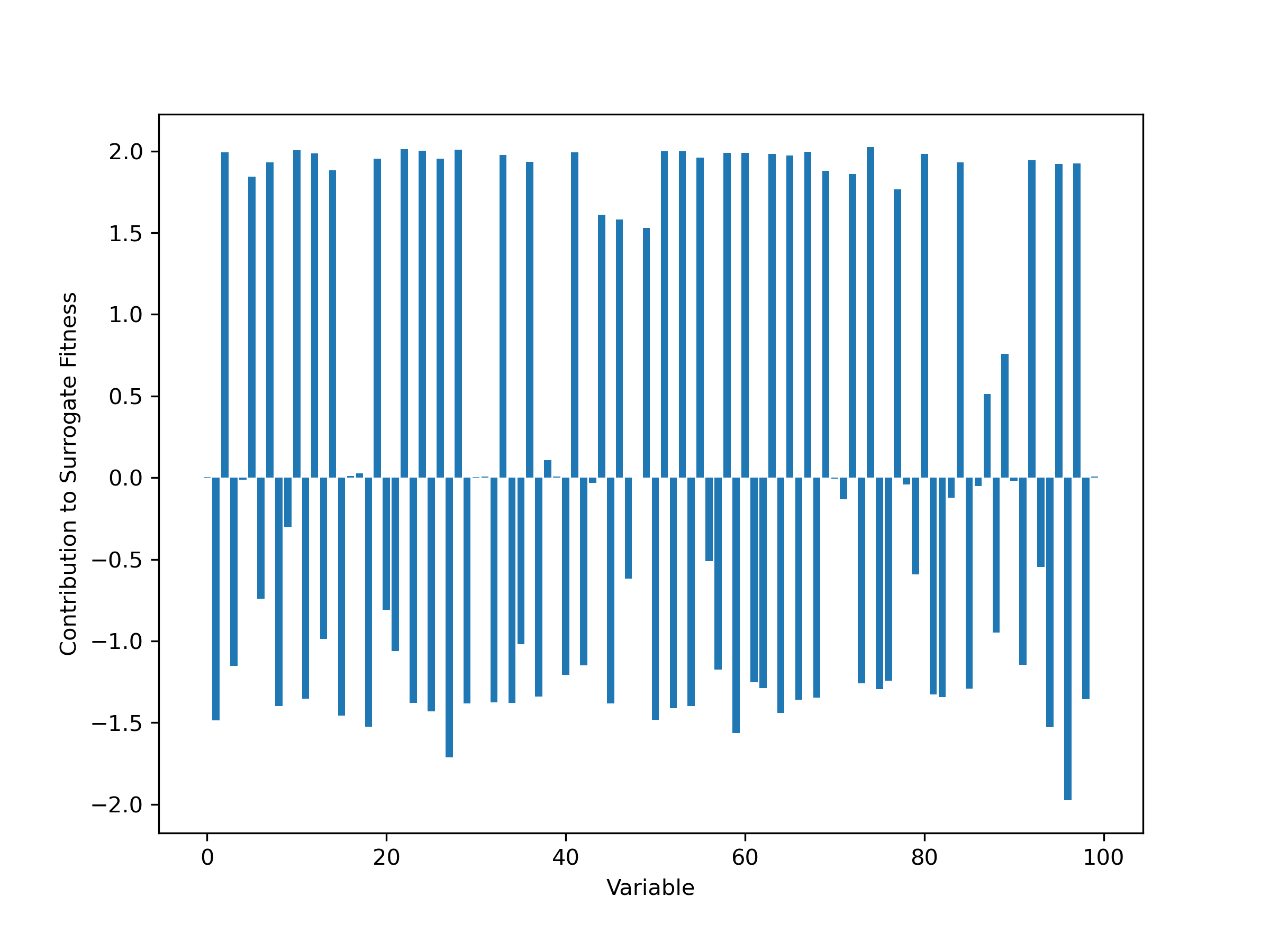}
  \caption{1D Checkerboard Contribution to Surrogate Fitness per Variable (All Generations)}
  \label{fig:1DCheckAll}
\end{figure}

\begin{figure}[htb]
  \centering
  \includegraphics[width=\linewidth]{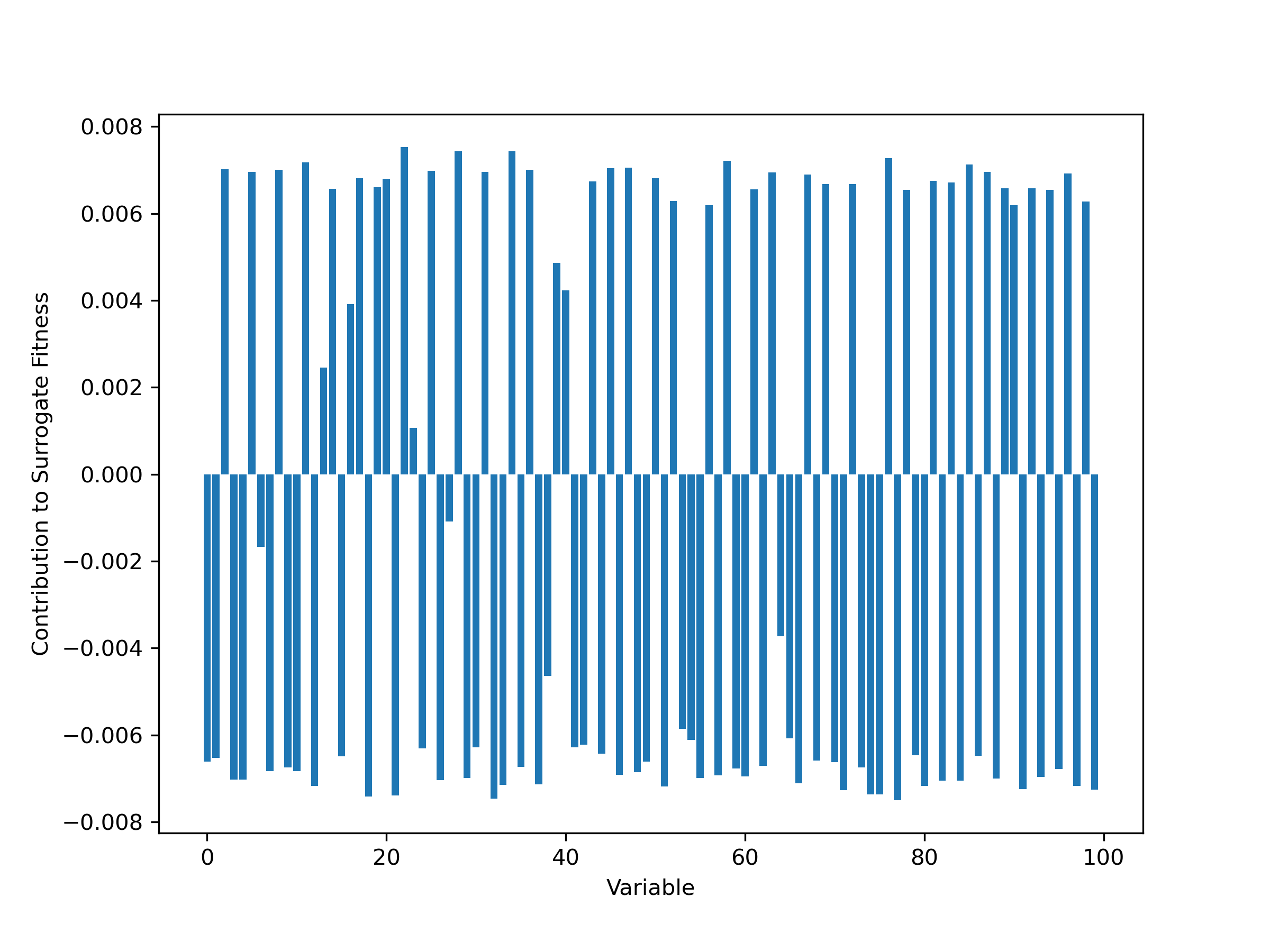}
  \caption{2D Checkerboard Contribution to Surrogate Fitness per Variable (All Generations)}
  \label{fig:2DCheckAll}
\end{figure}

\begin{figure}[htb]
  \centering
  \includegraphics[width=\linewidth]{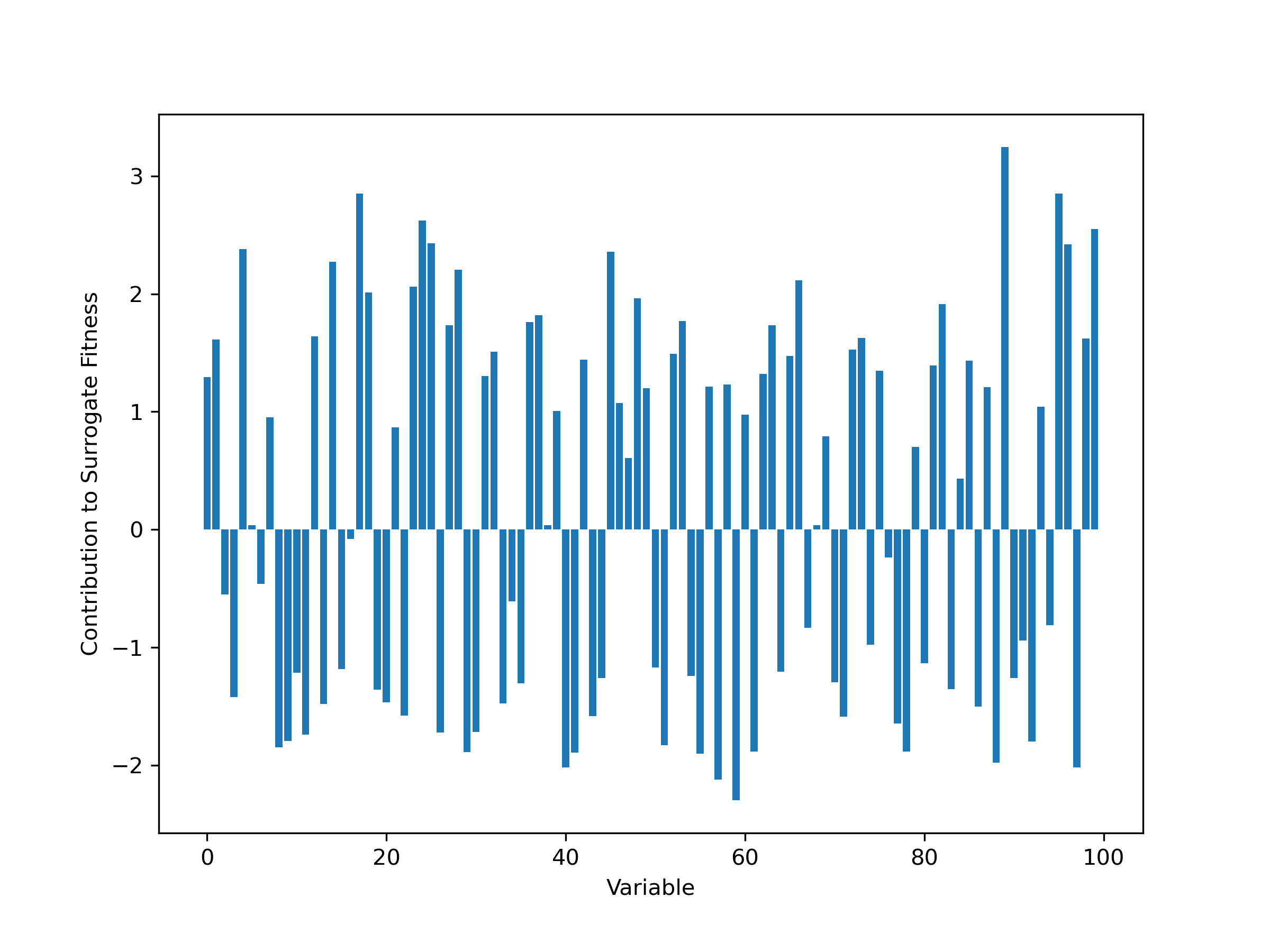}
  \caption{MAXSAT Contribution to Surrogate Fitness per Variable (All Generations)}
  \label{fig:MAXSATAll}
\end{figure}

\begin{figure}[htb]
  \centering
  \includegraphics[width=\linewidth]{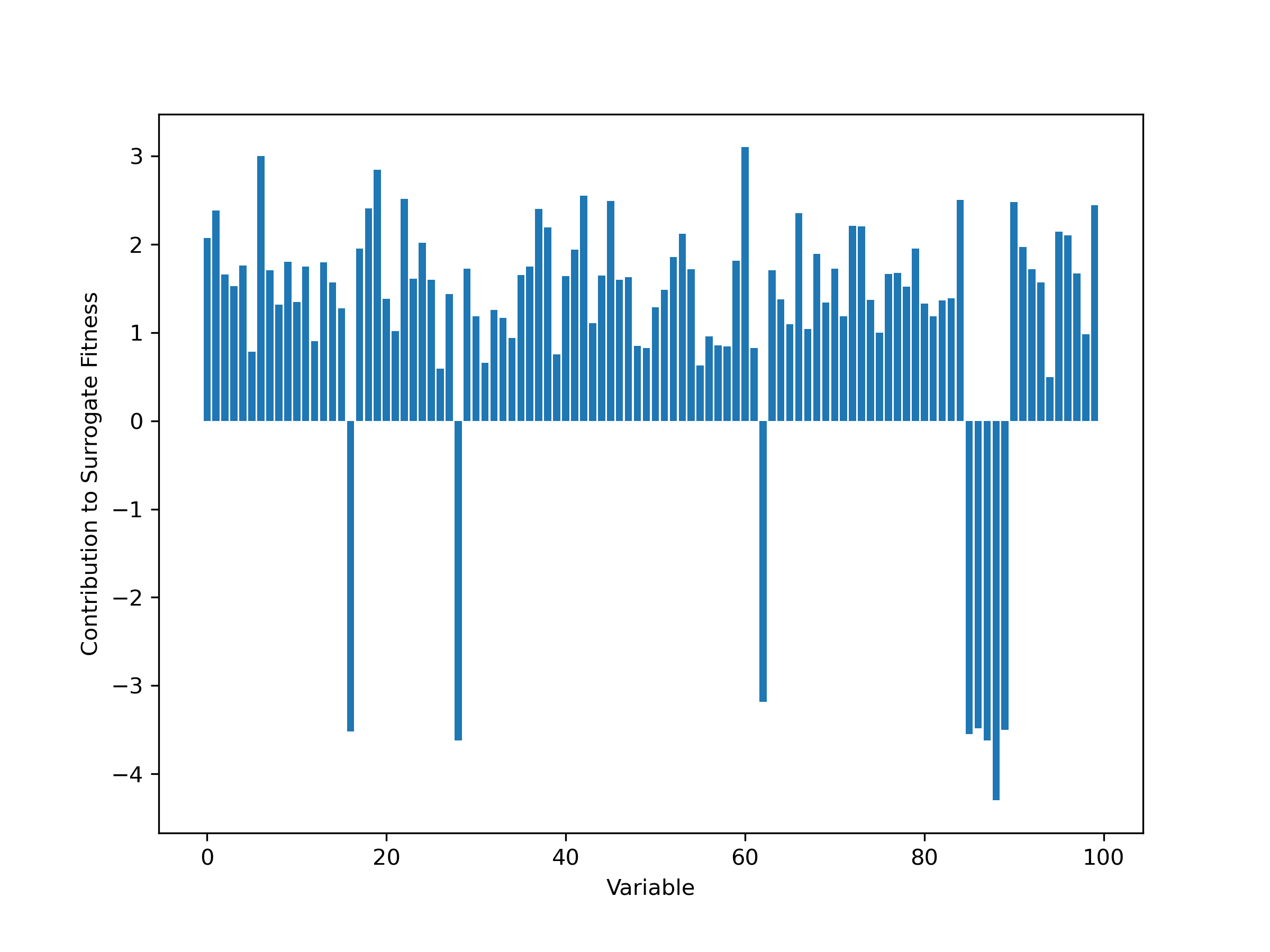}
  \caption{Trap5 Contribution to Surrogate Fitness per Variable (All Generations)}
  \label{fig:Trap5All}
\end{figure}

In this section we will consider the results of mining a surrogate fitted to all populations of the algorithm run. Each plot in  \Cref{fig:1DCheckAll,fig:2DCheckAll,fig:MAXSATAll,fig:Trap5All} shows the mean contribution to surrogate fitness for each variable. The positive values indicate that the optimal value was a 0 and negative values indicate that it was a 1. 

Comparing these plots to those generated for only the first population data, we see a marked contrast. We observe that the surrogate model has picked up key characteristics of the problem; for Trap5 we can now start to see clearer groupings of variables as we would expect for this problem. However the surrogate is reflecting that the GA has converged on a local optima, and hence the clear grouping that has settled on all 0 for variables 85-90.  

For the 1D Checkerboard we see a clearer pattern of alternating 1s and 0s and this pattern is more pronounced, and each variable seems to be contributing equally to fitness. 

For 2D Checkerboard, we see that the variables share a similar relationship to fitness and fall on or around the middle, and we see some regular groupings of slightly more dominant variables, which seem mostly clustered around the central region of our checkerboard.

For MAXSAT, we see similar grouping of clauses and clustering of the variables that contribute more to overall fitness.

We can note that the explanatory method of using a surrogate model does reveal that the model has begun to detect the key components of the problem.
% \clearpage

\vspace{-0.5cm}

\clearpage
\section{Acknowledgments}
The Data Lab funded project (Grant: 191047)

\section{Conclusion}\label{conclusion}

This paper set out to investigate the possibility of mining surrogate models for explanations about solution quality at different stages of an evolutionary run. We looked at the surrogate models learning of the search space for four benchmark problems; {\itshape 1D Checkerboard, 2D Checkerboard, MAXSAT and Trap5}. We did this through using the population and fitness values for individuals generated by a GA. We used this population data as input to train an SVR model, which we mined for a deeper understanding of the solutions generated by the GA for each generation.

This work has demonstrated a starting point for further exploration and more importantly the generation of explanations around the near-optimal solution presented by a GA. We focused on mining the surrogate to understand how the mean contribution of each variable changed as the GA generated more insights into the population and problem at hand. In particular, we compared a surrogate model trained on the initial population against one trained on population data after 100 generations. These experiments suggest that, for the purposes of explanation, it is preferable to train the surrogate on all solutions visited over the course of the GA run. Although, there is also some evidence to suggest that useful information about the problem can be mined during the first generation. Further investigation into what is possible at each stage of the run is thus needed.

Future work will involve analysis of interaction of variables and investigation of multi-modal problems. As understanding of this approach to explanations grows we will apply the above method to real-world problems. We will also further investigate the performance of a surrogate model at different stages of a GA run; to generate comparisons at different generations of the GA run and investigate at what point it would be beneficial to switch from using a costly fitness function to the surrogate model and comparison of solutions generated between the two scenarios. 

We also propose to further extend this work to exploit recent but already well known techniques for explaining machine learning models, such as permutation feature importance measurement\cite{breiman_random_2001, fisher_all_2018, wei_variable_2015} and SHapley Additive exPlanations (SHAP) \cite{lundberg_unified_2017} as a way to explain variable selection for individual solutions. The work presented here is a first step towards building a more generalised framework for explainability within the evolutionary computation (EC) domain.

The ultimate aim of this work is to generate explanations of the solution generated, by mining the surrogate model, and presenting these explanations to the end-user to instill more confidence and trust in our model. In turn, this increased trust in the solutions should lead to greater uptake and increased benefit from the use of GAs in practice.

%\clearpage

% \clearpage
\bibliographystyle{ACM-Reference-Format}
\bibliography{GECCO-2022-WS}
\end{document}